\documentclass[11pt]{article}

\usepackage[preprint]{acl}

\usepackage{times}
\usepackage{latexsym}
\usepackage{amsmath}
\usepackage{amssymb}
\usepackage{booktabs}
\usepackage{placeins}
\usepackage{xcolor}
\usepackage{multirow} 
\usepackage[table]{xcolor}
\definecolor{best}{RGB}{255, 204, 214}    
\definecolor{second}{RGB}{255, 235, 238}  
\usepackage[most]{tcolorbox}
\usepackage{enumitem}

\usepackage[T1]{fontenc}

\usepackage[utf8]{inputenc}

\usepackage{microtype}

\usepackage{inconsolata}

\usepackage{graphicx}

\usepackage[most]{tcolorbox}
\usepackage{setspace}
\tcbuselibrary{breakable}
\usepackage{listings}

\usepackage{algorithm}
\usepackage{algorithmic}

\title{Vital Trace: Protocol-Constrained Patient-State Reasoning for Longitudinal Clinical Trajectories}

\author{Zhan Qu and Michael Färber\\
  TU Dresden and ScaDS.AI, Germany \\
  \texttt{\{zhan.qu, michael.faerber\}@tu-dresden.de} \\}

\begin{document}
\maketitle

\begin{abstract}

Longitudinal clinical reasoning over electronic health records requires tracking evolving physiological measurements, laboratory results, and interventions across extended patient trajectories. Existing LLM-based clinical reasoning systems often rely on repeatedly serializing patient histories or exchanging unconstrained textual agent messages, leading to context drift, unstable reasoning, and growing inference cost over long horizons. We present \textbf{Vital Trace}, a protocol-constrained multi-agent framework for future clinical risk prediction over evolving ICU trajectories. Instead of maintaining unbounded textual histories, Vital Trace uses a compact persistent patient-state memory together with staged reasoning performed by four coordinated agents: a Router, Reasoner, Auditor, and Steward. To support temporally coherent reasoning, we introduce a manually curated Global Protocol containing physiological state-transition rules and a dynamic patient-state representation that tracks hemodynamic, respiratory, renal, metabolic, and inflammatory instability over time. We evaluate Vital Trace on MIMIC-IV and eICU using future vasopressor-support, respiratory-support, renal-support, and deterioration prediction tasks. Results show that structured protocol-constrained reasoning improves temporal consistency, communication stability, calibration, and interpretability compared with free-form multi-agent baselines while achieving strong predictive performance across long ICU trajectories.
\end{abstract}

\begin{figure*}[!ht]
    \centering
    \includegraphics[width=1.0\linewidth]{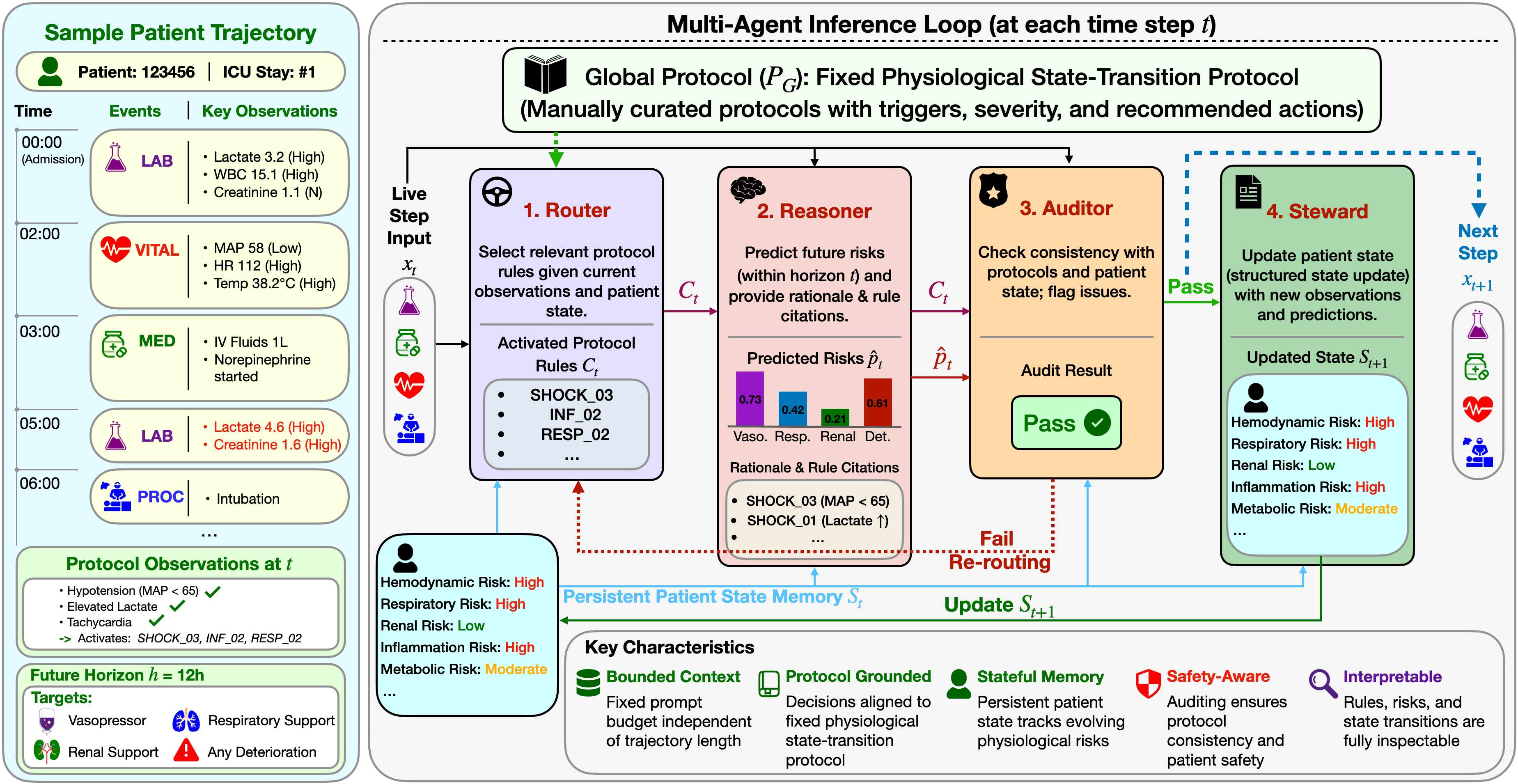}
    \vspace{-0.5cm}
    \caption{
    \textbf{Overview of Vital Trace.}
    Vital Trace performs protocol-grounded longitudinal ICU reasoning over clinical trajectories using a staged multi-agent inference loop. At each timestep \(t\), a structured clinical packet \(x_t\) is processed together with a fixed manually curated \emph{Global Protocol} \(\mathcal{P}_G\) and a persistent patient-state memory \(\mathcal{S}_t\). The \textbf{Router} activates clinically relevant protocol rules, the \textbf{Reasoner} predicts future intervention risks with protocol-grounded citations, the \textbf{Auditor} verifies protocol-state consistency and safety constraints, and the \textbf{Steward} updates the persistent physiological state \(\mathcal{S}_{t+1}\). By evolving structured patient states, Vital Trace maintains a bounded context during long-horizon reasoning. The left panel illustrates a representative ICU trajectory with protocol-triggering observations, while the bottom panel summarizes the core design properties of Vital Trace: protocol grounding, persistent state evolution, safety-aware auditing, and interpretable longitudinal reasoning.
    }
    \vspace{-0.3cm}
    \label{fig:overview}
\end{figure*}

\section{Introduction}

Large language models (LLMs) have demonstrated strong performance across a range of medical NLP tasks, including clinical question answering, report generation, summarization, and medical knowledge reasoning~\citep{singhal2023large,nori2023capabilities,qu2025medieval}. However, longitudinal clinical reasoning over ICU trajectories remains substantially more challenging. Real-world critical care unfolds as a temporally evolving stream of heterogeneous physiological observations, interventions, medications, and procedures with irregular timing and long-range dependencies~\citep{wornow2024context,xu2024ram}. In this setting, clinically relevant evidence may persist across many hours or days, requiring models to maintain coherent representations of evolving patient state rather than reason over isolated observations.

Existing approaches primarily address longitudinal reasoning through larger context windows, retrieval augmentation, or sequential representation learning. Long-context LLMs attempt to preserve more historical information but suffer from increasing computational cost, context degradation, and unstable long-range attention over extended trajectories~\citep{liu2024lost,press2022train,wornow2024context}. Retrieval-augmented systems can incorporate external medical knowledge, yet retrieved information may be temporally irrelevant or clinically inconsistent with the current patient state~\citep{xu2024ram,xia2025mmedrag}. Structured EHR foundation models improve temporal representation learning~\citep{waxler2025generative,poulain2024graph,cui2024automated}, but most approaches still treat longitudinal history as a flat sequence of events or tokens without explicit mechanisms for persistent patient-state evolution, protocol-grounded reasoning, or interpretable temporal state transitions.

We introduce \textbf{Vital Trace}, a protocol-constrained multi-agent framework for longitudinal ICU reasoning based on persistent physiological patient-state evolution. Instead of repeatedly serializing growing patient histories into free-form prompts, Vital Trace maintains a compact persistent patient-state memory and performs reasoning through staged structured communication. The framework consists of four coordinated agents: a \emph{Router} that activates clinically relevant protocol rules, a \emph{Reasoner} that predicts future intervention risks, an \emph{Auditor} that verifies protocol consistency and safety constraints, and a \emph{Steward} that updates persistent physiological state over time.

Vital Trace separates memory into two complementary components: (i) a fixed manually curated \emph{Global Protocol} containing clinically grounded temporal state-transition rules, and (ii) a dynamic persistent patient-state memory that tracks evolving physiological instability across longitudinal ICU trajectories. The Global Protocol acts as a stabilizing temporal prior for staged reasoning, while persistent patient-state memory enables compact longitudinal state tracking without unbounded prompt growth. Rather than replacing neural reasoning with symbolic rules, protocol constraints guide structured inter-agent communication and interpretable patient-state evolution.

We evaluate Vital Trace on longitudinal ICU trajectories derived from MIMIC-IV and eICU using future intervention prediction tasks including vasopressor support, respiratory support, renal support, and overall deterioration. Beyond standard predictive metrics, we evaluate temporal early-warning performance, calibration, protocol consistency, communication stability, and counterfactual consistency under clinically meaningful physiological perturbations. Experimental results demonstrate that structured patient-state communication improves longitudinal reasoning stability, protocol consistency, and interpretability compared to free-form multi-agent reasoning while maintaining competitive predictive performance over extended ICU trajectories. 
Our major contributions are summarized as follows:
\begin{itemize}

    \item We introduce \textbf{Vital Trace}, a protocol-constrained multi-agent framework for longitudinal reasoning that performs future risk prediction using persistent physiological state tracking over evolving patient trajectories.

    \item We propose a transition-centered clinical representation that organizes ICU trajectories around physiological state changes and temporal trends rather than flat event serialization or continuously expanding textual histories.

    \item We introduce a dual-memory reasoning architecture consisting of (i) a fixed manually curated Global Protocol of physiological state-transition rules and (ii) a dynamic patient-state memory that enables compact and interpretable longitudinal reasoning across extended ICU trajectories.

    \item We develop a structured inference loop with protocol-guided routing, audit-based consistency checking, and constrained patient-state updates to stabilize multi-agent reasoning over long clinical trajectories.

    \item We perform comprehensive evaluation on MIMIC-IV and eICU using future intervention and deterioration prediction tasks, including analyses of predictive performance, calibration, protocol consistency, temporal stability, and counterfactual robustness.

\end{itemize}

\section{Related Work}
\label{sec:related}

\textbf{Medical Foundation Models and Longitudinal EHR Reasoning.}
Recent medical foundation models have demonstrated strong performance on clinical reasoning, prediction, and generation tasks using large-scale structured and unstructured medical data \citep{singhal2023large,nori2023capabilities,wornow2023shaky,chen2023meditron}. Beyond text-centric medical LLMs, several works model longitudinal EHR trajectories directly, including COMET \citep{waxler2025generative}, GT-BEHRT \citep{poulain2024graph}, and AutoDP \citep{cui2024automated}, which improve temporal representation learning through graph modeling, sequential pretraining, or task-aware architectures. Other approaches investigate long-context patient modeling \citep{wornow2024context}, retrieval-augmented clinical systems \citep{xu2024ram,xia2025mmedrag}, and generative longitudinal EHR modeling \citep{he2024flexible,zhong2024synthesizing}. In contrast to these approaches, Vital Trace focuses on persistent patient-state reasoning and protocol-grounded physiological state evolution rather than scaling context length, retrieval capacity, or purely sequential representation learning.

\noindent \textbf{Memory-Augmented and Agentic Reasoning Systems.}
Our work is also related to recent research on memory-augmented and agentic LLM systems that maintain evolving inference state without parameter updates \citep{madaan2023self,shinn2023reflexion,packer2023memgpt,suzgun2025dynamic,zhang2025agentic}. These approaches typically rely on iterative reflection, free-form textual memory, or unconstrained inter-agent communication to improve long-horizon reasoning behavior. In contrast, Vital Trace maintains a compact protocol-grounded patient-state representation designed specifically for longitudinal physiological reasoning, safety-constrained inference, and interpretable patient-state evolution in ICU trajectories.

\section{Methodology}
\label{sec:method}

We formulate longitudinal ICU reasoning as online risk estimation over a patient trajectory
\(
\mathcal{X}=(x_1,\ldots,x_T)
\),
where each \(x_t\) denotes a longitudinal clinical packet derived from temporally evolving physiological observations and intervention events. At each timestep, the system observes \(x_t\), predicts future clinical risks within a fixed prediction horizon \(h\), and updates persistent patient-state memory before processing the next packet. The prediction target is a multi-label risk vector
\begin{equation}
\mathbf{y}^{(h)}_t =
[
y^{\mathrm{vaso}}_t,
y^{\mathrm{resp}}_t,
y^{\mathrm{renal}}_t,
y^{\mathrm{det}}_t
]
\in \{0,1\}^{4}
\label{eq:targets}
\end{equation}
corresponding to future vasopressor support, respiratory support, renal support, and overall deterioration.

Rather than repeatedly serializing the growing history \(x_{1:t}\) into free-form prompts, Vital Trace performs protocol-constrained patient-state reasoning over a compact persistent inference state
\begin{equation}
\mathcal{M}_t = (\mathcal{P}_G, \mathcal{S}_t)
\label{eq:memory}
\end{equation}
where \(\mathcal{P}_G\) is a fixed Global Protocol shared across patients and \(\mathcal{S}_t\) is a dynamic patient-state memory tracking evolving physiological instability. At each timestep, a staged Router--Reasoner--Auditor--Steward loop operates over \((x_t,\mathcal{M}_t)\) to produce calibrated intervention risks and an updated state \(\mathcal{S}_{t+1}\). This design maintains bounded inference context while preserving clinically meaningful longitudinal information.

\subsection{Dual Memory for Longitudinal Reasoning}
\label{sec:memory}

Longitudinal reasoning memory is decomposed into two complementary components: a fixed global protocol and a dynamic patient-state.

\paragraph{Global Protocol.}

The Global Protocol
\(
\mathcal{P}_G=\{r_j\}_{j=1}^{N}
\)
is a compact manually curated rule library designed for longitudinal ICU deterioration reasoning. Each protocol rule specifies:
(i) temporal trigger conditions,
(ii) physiological state transitions,
(iii) severity levels,
(iv) linked prediction targets,
and
(v) counterfactual intervention candidates.

The trigger field defines clinically relevant physiological patterns such as persistent hypotension, rising lactate, worsening creatinine, hypoxemia, inflammatory escalation, or recovery dynamics. The transition field specifies how activated rules modify persistent physiological state dimensions, while target mappings associate rules with downstream prediction tasks.

The protocol is not intended to encode complete ICU guidelines. Instead, it provides a high-coverage set of deterioration and recovery patterns aligned with longitudinal intervention prediction over MIMIC-IV and eICU. Rules are grouped into hemodynamic, respiratory, renal, inflammatory, metabolic, recovery, and multi-domain deterioration categories. Detailed protocol templates and representative rules are provided in Appendix~\ref{app:global_protocol}.

\paragraph{Patient-State Memory.}

The patient-state
\begin{equation}
\mathcal{S}_t =
[
s^{\mathrm{hemo}}_t,
s^{\mathrm{resp}}_t,
s^{\mathrm{renal}}_t,
s^{\mathrm{metabolic}}_t,
s^{\mathrm{inflam}}_t
]
\label{eq:patient_state}
\end{equation}
represents persistent physiological instability across five longitudinal risk dimensions: hemodynamic, respiratory, renal, metabolic, and inflammatory state. Rather than storing free-form textual summaries, each dimension maintains an evolving structured risk state updated throughout the trajectory. The metabolic and inflammatory dimensions primarily contribute to generalized deterioration reasoning rather than defining independent intervention-specific prediction targets.

In implementation, state variables are represented as bounded discrete scores updated according to current physiological evidence, activated protocol rules, and audit outcomes. This structured state serves as a compact interface between historical trajectory evidence and future reasoning, preserving clinically meaningful longitudinal dynamics without context growth proportional to trajectory length.

\subsection{Longitudinal Clinical Representation}
\label{sec:packet_representation}

Vital Trace operates over longitudinal clinical packets constructed around clinically meaningful physiological state transitions rather than fixed temporal windows or flat event serialization. Each packet aggregates:
(i) normalized physiological measurements,
(ii) abnormality indicators,
(iii) intervention signals,
(iv) temporal trend descriptors,
and
(v) protocol-relevant symbolic observations.

Examples include persistent hypotension, rising lactate, worsening creatinine, oxygenation decline, inflammatory escalation, and sustained recovery patterns. By organizing trajectories around physiological transitions rather than raw event density, the representation preserves clinically meaningful temporal structure while reducing noise from redundant high-frequency measurements.

This transition-centered representation additionally stabilizes inter-agent communication by providing compact protocol-ready contexts instead of unconstrained free-form trajectory histories.

\subsection{Protocol-Constrained Inference Loop}
\label{sec:architecture}

At each timestep \(t\), Vital Trace executes four sequential agents operating over the same pretrained backbone model within a given experiment.

\paragraph{1. Router (protocol activation).}

The Router activates a compact subset of protocol rules relevant to the current clinical packet:
\begin{equation}
R_t = \mathrm{Router}(x_t,\mathcal{P}_G)
\label{eq:routing}
\end{equation}
Routing is trigger-guided; the Router matches physiological observations and temporal trends in \(x_t\) to protocol triggers in \(\mathcal{P}_G\), producing a bounded set of evidence-supported rule activations for downstream reasoning. This stage acts as symbolic attention over protocol memory and prevents uncontrolled context expansion during long-horizon inference.

\paragraph{2. Reasoner (future risk prediction).}

The Reasoner predicts future intervention risks conditioned on the current packet, activated protocol rules, and patient-state memory:
\begin{equation}
\hat{\mathbf{p}}_t =
\mathrm{Reasoner}(x_t,R_t,\mathcal{S}_t),
\qquad
\hat{\mathbf{p}}_t \in [0,1]^4
\label{eq:reasoner}
\end{equation}

The output includes risk probabilities, predicted intervention bundles, protocol citations, and optional counterfactual notes. Protocol citations are constrained to activated rules in \(R_t\), ensuring that generated reasoning remains grounded in protocol-supported physiological evidence.

\paragraph{3. Auditor (protocol-state consistency checking).}

The Auditor evaluates whether Reasoner outputs are consistent with activated protocol rules and the current patient state:
\begin{equation}
a_t =
\mathrm{Auditor}(\hat{\mathbf{p}}_t,R_t,\mathcal{S}_t,x_t)
\label{eq:auditor}
\end{equation}

The audit output contains PASS/FAIL decisions together with structured issue tags including unsupported predictions, missing active-rule risks, protocol contradictions, and state-transition inconsistencies. Final adjudication is deterministic, while LLM-generated audit outputs are retained as auxiliary evidence for debugging and interpretability analysis. If the audit fails, structured corrective feedback is returned to the Router and Reasoner, triggering constrained re-routing and re-sampling. To prevent infinite correction loops, Vital Trace applies a conservative fallback policy after two consecutive audit failures, producing bounded-risk predictions using only high-confidence protocol-supported outputs.

\paragraph{4. Steward (persistent state evolution).}

The Steward updates patient-state memory after prediction and auditing:
\begin{equation}
\mathcal{S}_{t+1} =
\mathrm{Steward}(
\mathcal{S}_t,
x_t,
R_t,
\hat{\mathbf{p}}_t,
a_t
)
\label{eq:steward}
\end{equation}

State updates are constrained by activated protocol transitions and bounded state ranges. For example, repeated hypotension with rising lactate may increase hemodynamic instability, while sustained recovery with improving perfusion may decrease it. If the audit fails, upward risk-state transitions are performed conservatively. Unlike free-form summarization, the Steward performs explicit physiological state transitions over predefined instability dimensions, producing an interpretable longitudinal reasoning trace across ICU trajectories.

\subsection{Inference-Time Operation}
\label{sec:inference_regime}

Vital Trace is an inference-time orchestration framework operating over frozen pretrained LLMs. No backbone parameters are fine-tuned during staged inference. Behavioral differences arise from four design choices:
(i) transition-based longitudinal packet construction,
(ii) protocol-constrained routing,
(iii) structured role decomposition,
and
(iv) persistent patient-state evolution.

The online prediction process is:
\[
(x_t,\mathcal{P}_G,\mathcal{S}_t)
\rightarrow
R_t
\rightarrow
\hat{\mathbf{p}}_t
\rightarrow
a_t
\rightarrow
\mathcal{S}_{t+1}.
\]

Only \(\mathcal{S}_t\) evolves across timesteps; the Global Protocol remains fixed throughout evaluation. This prevents test-time protocol leakage while allowing patient-specific adaptation through persistent physiological state evolution. Because transmitted memory has fixed structure, the inference context remains bounded with respect to trajectory length.

\section{Data Construction}
\label{sec:data}

We construct longitudinal ICU trajectories centered around clinically meaningful physiological state transitions rather than fixed temporal windows or flat event serialization. Raw EHR records are heterogeneous, irregularly sampled, and distributed across multiple relational tables containing physiological measurements, medications, procedures, and intervention events. Vital Trace transforms these records into compact transition-aware longitudinal clinical packets aligned with persistent patient-state reasoning.

\textbf{Source Datasets.}
We evaluate on MIMIC-IV v2.2~\citep{johnson2023mimic} and eICU~\citep{PhysioNet-eicu-crd-2.0}, two large-scale ICU datasets containing longitudinal physiological measurements, interventions, and clinical events. For each ICU stay, we extract laboratory events, bedside observations (vitals), and interventions (medications and procedures), from which intervention targets are derived. 

\textbf{Canonical Signal Construction.}
All extracted records are mapped into harmonized physiological signal representations containing normalized measurements, units, abnormality indicators, and temporal metadata. Feature aliases across datasets are unified into shared physiological concepts including hemodynamic, respiratory, renal, inflammatory, and metabolic signals.

Abnormality indicators are derived using a hybrid scheme: source-provided abnormal/reference annotations when available, and manually curated protocol thresholds when source flags are unavailable or not directly comparable. The preprocessing pipeline additionally derives temporal trend indicators including rising lactate, worsening creatinine, persistent hypotension, oxygenation decline, and recovery transitions.

\textbf{Transition-Based Longitudinal Packets.}
Rather than partitioning trajectories into fixed temporal windows, Vital Trace constructs longitudinal packets around clinically meaningful signal-state transitions. A new packet is created when tracked physiological signals undergo clinically relevant transitions, including:
(i) normal-to-abnormal changes,
(ii) abnormal-to-normal recovery,
(iii) severity-level transitions,
or
(iv) trend-state transitions such as stable-to-rising lactate.

Within each packet, representative physiological measurements and transition descriptors are retained together with intervention signals and temporal metadata. This transition-centered representation preserves clinically meaningful trajectory structure while reducing noise from redundant high-frequency measurements and avoiding excessive trajectory fragmentation.

\textbf{Intervention Target Construction.}
Future-horizon labels are generated using hierarchical intervention concept mapping over medications, procedures, and intervention events. Mapping prioritizes:
(i) exact code matching,
(ii) normalized-code matching,
(iii) description-level mappings,
and
(iv) regex-based fallback matching.

We evaluate four prediction targets:
\begin{equation}
\mathbf{y}_t
=
[
y_t^{\mathrm{vaso}},
y_t^{\mathrm{resp}},
y_t^{\mathrm{renal}},
y_t^{\mathrm{det}}
]
\label{eq:prediction_targets}
\end{equation}
corresponding to future vasopressor support signal, respiratory support signal, renal support signal, and overall deterioration.

\textbf{Trajectory Filtering and Serialization.}
The preprocessing pipeline applies staged cohort filtering, signal-consistency filtering, transition validation, and intervention-mapping verification before serialization. Final trajectories contain longitudinal clinical packets, protocol-trigger observations, intervention labels, and metadata required for staged reasoning and counterfactual evaluation. Additional preprocessing details, dataset statistics, and examples are provided in Appendix~\ref{app:experiments}.

\section{Experiments}
\label{sec:experiments}

\subsection{Experimental Setup}
\label{sec:experimental_setup}

We evaluate Vital Trace using a streaming predict-then-update protocol over longitudinal ICU trajectories. At each timestep \(t\), the system observes the current longitudinal clinical packet \(x_t\), predicts future intervention risks within a fixed prediction horizon \(h\), and updates persistent patient-state memory before processing the next packet. This setting prevents temporal leakage and reflects realistic online clinical deployment.

\textbf{Datasets and Statistics.}
Experiments are conducted on longitudinal ICU trajectories derived from MIMIC-IV and eICU. We intentionally construct a smaller but high-quality longitudinal cohort to enable detailed protocol-grounded trajectory analysis and counterfactual evaluation. After preprocessing, the MIMIC-IV cohort contains \texttt{1,000} ICU stays (\texttt{1,000} trajectories) and \texttt{80,064} longitudinal clinical packets, while the eICU cohort contains \texttt{1,000} ICU stays (\texttt{1,000} trajectories) and \texttt{75,596} packets. The average number of packets per stay is \texttt{80.1} for MIMIC-IV and \texttt{75.6} for eICU. The processed longitudinal trajectory dataset will be made publicly available through PhysioNet. Positive-label prevalence for vasopressor-support, respiratory-support, renal-support, and deterioration prediction tasks is \texttt{21.7\%}, \texttt{5.7\%}, \texttt{6.4\%}, and \texttt{30.2\%} on MIMIC-IV, and \texttt{52.2\%}, \texttt{35.2\%}, \texttt{9.1\%}, and \texttt{65.2\%} on eICU, respectively.



\textbf{Backbone Models and Baselines.}
We evaluate Vital Trace using 7 open-source instruction-tuned LLM backbones spanning different model families, scales, and biomedical specialization levels. Evaluated models include GPT \citep{agarwal2025gpt}, LLaMA \citep{grattafiori2024llama}, Mistral \citep{jiang2024mixtralexperts}, Qwen \citep{yang2025qwen3technicalreport}, and 2 biomedical models, Meditron \citep{chen2023meditron} and ClinicalCamel \citep{toma2023clinical}. All agents within a given experiment share the same backbone model, enabling controlled comparison of protocol-grounded longitudinal reasoning across model families and parameter scales. We compare Vital Trace against:
(i) single-agent prompting baselines,
(ii) free-form multi-agent systems using unrestricted textual communication,
and
(iii) structured ablations removing protocol routing, auditing, or persistent patient-state evolution.

\textbf{Evaluation Metrics.}
Predictive performance is evaluated using Macro-AUROC, Macro-AUPRC, Micro-F1, and Micro-Expected Calibration Error (Micro-ECE). For each trajectory, we construct intervention-style physiological perturbations on protocol-relevant variables while preserving the surrounding temporal context. We then evaluate whether predicted risks, activated protocol rules, and patient-state transitions change in clinically consistent directions. Additional evaluation metrics and results are provided in Appendix~\ref{app:additional_results}.

\textbf{Implementation Details.}
Inference is performed under a fixed context budget using longitudinal clinical packets and structured protocol-grounded communication. The Global Protocol remains fixed throughout all experiments, while persistent patient-state memory evolves dynamically through Steward updates. All agents operate through structured JSON-based communication without parameter updates during inference. Additional implementation details, global protocol specifications, and prompt templates are provided in Appendix \ref{app:experiments}, \ref{app:global_protocol}, and \ref{app:prompts}.

\begin{table*}[!ht]
\caption{Main results on MIMIC-IV and eICU (Values in brackets denote bootstrapped 95\% confidence intervals).}
\vspace{-0.3cm}
\resizebox{\textwidth}{!}{
\begin{tabular}{l|cccc|cccc}
\toprule
\multirow{2}{*}{Method} & \multicolumn{4}{c|}{MIMIC-IV} & \multicolumn{4}{c}{eICU} \\
& Macro-AUROC $\uparrow$ & Macro-AUPRC $\uparrow$ & Micro-F1 $\uparrow$ & Micro-ECE $\downarrow$ & Macro-AUROC $\uparrow$ & Macro-AUPRC $\uparrow$ & Micro-F1 $\uparrow$ & Micro-ECE $\downarrow$ \\
\midrule
\textit{Single-Agent LLM} \\
gpt-oss-20b & 0.472 [0.423, 0.656] & 0.269 [0.207, 0.552] & 0.349 & 0.359 & 0.468 [0.289, 0.614] & 0.299 [0.243, 0.409] & 0.304 & 0.249 \\
Llama-3.1-8B-Instruct & 0.539 [0.423, 0.619] & 0.319 [0.186, 0.428] & 0.345 & 0.530 & 0.461 [0.298, 0.675] & 0.274 [0.238, 0.447] & 0.423 & 0.273 \\
Llama-3.3-70B-Instruct & 0.614 [0.413, 0.776] & 0.376 [0.245, 0.514] & 0.406 & 0.267 & 0.487 [0.174, 0.606] & 0.351 [0.115, 0.431] & 0.373 & 0.317 \\
Meditron-70B & 0.463 [0.221, 0.489] & 0.253 [0.193, 0.324] & 0.382 & 0.442 & 0.491 [0.312, 0.601] & 0.353 [0.217, 0.517] & 0.411 & 0.304 \\
\midrule
\textit{Free-Form Multi-Agent} \\
gpt-oss-20b & 0.493 [0.376, 0.606] & 0.262 [0.195, 0.467] & 0.378 & 0.315 & 0.517 [0.327, 0.739] & 0.381 [0.199, 0.603] & 0.399 & 0.373 \\
Llama-3.1-8B-Instruct & 0.562 [0.333, 0.654] & 0.336 [0.194, 0.404] & 0.387 & 0.384 & 0.537 [0.358, 0.619] & 0.397 [0.391, 0.505] & 0.401 & 0.298 \\
Llama-3.3-70B-Instruct & \cellcolor{second} 0.768 [0.517, 0.827] & \cellcolor{second} 0.459 [0.246, 0.496] & 0.416 & 0.274 & 0.535 [0.371, 0.583] & 0.397 [0.161, 0.517] & 0.370 & 0.324 \\
Meditron-70B & 0.526 [0.291, 0.601] & 0.320 [0.198, 0.431] & 0.347 & 0.289 & 0.514 [0.362, 0.574] & 0.380 [0.218, 0.617] & 0.339 & 0.316 \\
\midrule
\textit{Vital Trace} \\
Llama-3.1-8B-Instruct & 0.656 [0.544, 0.835] & 0.321 [0.209, 0.437] & 0.463 & 0.337 & 0.529 [0.496, 0.622] & 0.368 [0.181, 0.695] & 0.478 & 0.242 \\
Llama-3.3-70B-Instruct & \cellcolor{best} 0.834 [0.739, 0.935] & \cellcolor{best} 0.502 [0.423, 0.658] & \cellcolor{best} 0.622 & \cellcolor{best} 0.089 & \cellcolor{best} 0.637 [0.442, 0.836] & 0.393 [0.211, 0.543] & \cellcolor{best} 0.679 & \cellcolor{best} 0.116 \\
gpt-oss-20b & 0.604 [0.476, 0.636] & 0.308 [0.195, 0.467] & 0.378 & 0.315 & 0.534 [0.412, 0.613] & 0.400 [0.323, 0.491] & 0.452 & 0.221 \\
Qwen3-32B & 0.623 [0.322, 0.713] & 0.338 [0.192, 0.511] & 0.397 & \cellcolor{second} 0.218 & 0.547 [0.306, 0.694] & \cellcolor{best} 0.412 [0.189, 0.515] & 0.471 & 0.183 \\
Mixtral-8x7B-Instruct-v0.1 & 0.566 [0.434, 0.723] & 0.359 [0.191, 0.542] & \cellcolor{second} 0.483 & 0.289 & 0.525 [0.401, 0.618] & 0.391 [0.372, 0.492] & 0.517 & 0.241 \\
Meditron-70B & 0.613 [0.448, 0.686] & 0.316 [0.177, 0.406] & 0.374 & 0.240 & \cellcolor{second} 0.553 [0.438, 0.598] & 0.406 [0.373, 0.511] & \cellcolor{second} 0.561 & \cellcolor{second} 0.171 \\
ClinicalCamel-70B & 0.625 [0.477, 0.637] & 0.317 [0.196, 0.497] & 0.389 & 0.295 & 0.544 [0.456, 0.611] & \cellcolor{second} 0.409 [0.368, 0.501] & 0.560 & 0.243 \\
\bottomrule
\end{tabular}
}
\vspace{-0.2cm}
\label{tab:main_performance_expanded}
\end{table*}

\begin{table*}[!ht]
\centering
\caption{Per-target predictive performance on MIMIC-IV using Llama-3.3-70B-Instruct.}
\vspace{-0.3cm}
\small
\resizebox{0.95\textwidth}{!}{
\begin{tabular}{l|cc|cc|cc|cc}
\toprule
\multirow{2}{*}{Method} &
\multicolumn{2}{c|}{Vasopressor} &
\multicolumn{2}{c|}{Resp.\ support} &
\multicolumn{2}{c|}{Renal support} &
\multicolumn{2}{c}{Deterioration} \\
& AUROC $\uparrow$ & AUPRC $\uparrow$
& AUROC $\uparrow$ & AUPRC $\uparrow$
& AUROC $\uparrow$ & AUPRC $\uparrow$
& AUROC $\uparrow$ & AUPRC $\uparrow$ \\
\midrule
\textit{Single-Agent LLM} & 0.648 & 0.444 & 0.466 & 0.129 & 0.762 & \textbf{0.436} & 0.579 & 0.495 \\
\textit{Free-Form Multi-Agent} & 0.830 & 0.604 & 0.730 & 0.312 & \textbf{0.866} & 0.421 & 0.647 & 0.498 \\
\textit{Vital Trace} & \textbf{0.837} & \textbf{0.622} & \textbf{0.855} & \textbf{0.345} & 0.863 & \textbf{0.436} & \textbf{0.781} & \textbf{0.603} \\
\bottomrule
\end{tabular}
}
\vspace{-0.3cm}
\label{tab:per_label_mimic}
\end{table*}

\subsection{Results}
\label{sec:results}

Table~\ref{tab:main_performance_expanded} summarizes predictive performance on MIMIC-IV and eICU across single-agent LLMs, free-form multi-agent systems, and Vital Trace. The evaluation setting is challenging: long trajectories (around 200 packets/stay), clinically heterogeneous, and highly imbalanced across several intervention targets, particularly respiratory-support prediction (\(5.7\%\) prevalence on MIMIC-IV). In this setting, Vital Trace consistently improves both predictive discrimination and calibration relative to single-agent and unrestricted multi-agent baselines.

The strongest overall performance is achieved by Vital Trace with Llama-3.3-70B-Instruct, reaching \(0.834\) Macro-AUROC and \(0.502\) Macro-AUPRC on MIMIC-IV while substantially reducing calibration error (ECE \(=0.089\)). On eICU, the same configuration achieves the best AUROC, Micro-F1, and calibration among all models despite a substantial dataset shift in intervention prevalence and trajectory structure. In particular, eICU contains shorter but substantially denser intervention trajectories, with deterioration prevalence increasing from \(30.2\%\) on MIMIC-IV to \(65.2\%\) on eICU.

Free-form multi-agent reasoning consistently improves ranking performance over single-agent prompting, especially for larger backbones. However, Vital Trace further improves both ranking quality and thresholded operating behavior, suggesting that structured patient-state evolution stabilizes reasoning over long trajectories. The gains are especially pronounced for smaller models: for example, Vital Trace improves the Micro-F1 of Llama-3.1-8B-Instruct from \(0.345\) to \(0.463\) on MIMIC-IV and from \(0.423\) to \(0.478\) on eICU. Notably, free-form multi-agent reasoning with Llama-3.3-70B-Instruct remains a strong ranking baseline (\(0.768\) AUROC on MIMIC-IV), indicating that the gains of Vital Trace arise not merely from multi-agent decomposition, but from protocol-guided longitudinal state evolution.

\noindent \textbf{Per-target Predictions.}
Table~\ref{tab:per_label_mimic} reports per-target performance on MIMIC-IV using Llama-3.3-70B-Instruct. Vital Trace achieves the strongest performance on vasopressor, respiratory-support, and generalized deterioration prediction while remaining competitive on renal-support prediction. The largest improvement is observed for respiratory-support prediction, where AUROC improves from \(0.730\) to \(0.855\) relative to free-form multi-agent reasoning. This task requires integrating evolving respiratory dynamics over long temporal horizons rather than relying on isolated observations, which likely explains the larger benefit of persistent patient-state evolution. Vital Trace also substantially improves generalized deterioration prediction (\(0.781\) AUROC and \(0.603\) AUPRC), consistent with its ability to accumulate multi-system physiological instability over time.

\begin{table}[!t]
\caption{Ablations on MIMIC-IV (Llama-3.3-70B).}
\vspace{-0.3cm}
\centering
\small
\resizebox{0.49\textwidth}{!}{
\begin{tabular}{lccc}
\toprule
Variant & AUROC $\uparrow$ & AUPRC $\uparrow$ & ECE $\downarrow$ \\
\midrule
\textbf{Full Vital Trace} & \textbf{0.834 [0.739, 0.935]} & \textbf{0.502 [0.423, 0.658]} & \textbf{0.089} \\
w/o Router & 0.767 [0.513, 0.790] & 0.333 [0.262, 0.576] & 0.238 \\
w/o Auditor & 0.752 [0.500, 0.769] & 0.294 [0.262, 0.509] & 0.239 \\
w/o Patient-State Memory & 0.759 [0.526, 0.760] & 0.394 [0.364, 0.579] & 0.233 \\
w/o Global Protocol & 0.762 [0.539, 0.769] & 0.298 [0.266, 0.576] & 0.236 \\
\bottomrule
\end{tabular}
}
\vspace{-0.6cm}
\label{tab:ablations_main}
\end{table}

\begin{figure*}[!ht]
    \centering
    \includegraphics[width=\textwidth]{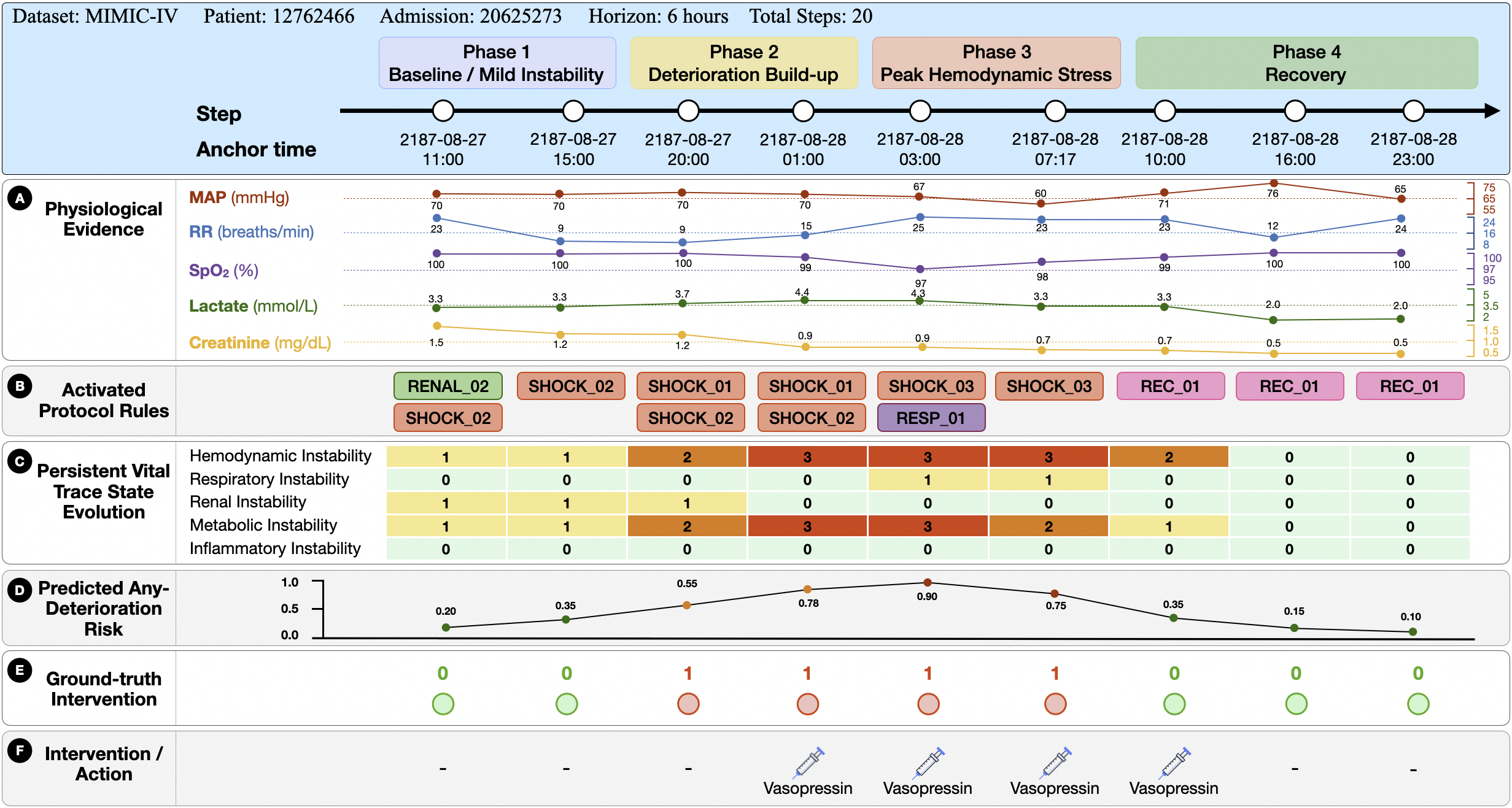}
    \vspace{-0.6cm}
    \caption{
    \textbf{Qualitative longitudinal reasoning example from Vital Trace on a MIMIC-IV trajectory using Llama-3.3-70B-Instruct.}
    The figure illustrates protocol-grounded longitudinal reasoning over a 6-hour prediction horizon. 
    \textbf{(A)} Temporal evolution of physiological evidence, including mean arterial pressure (MAP), respiratory rate (RR), oxygen saturation (SpO\(_2\)), lactate, and creatinine. Signal axes are centered around clinically relevant ranges (e.g., MAP \(<65\) mmHg, RR \(>24\) breaths/min, lactate \(>4\) mmol/L). 
    \textbf{(B)} Global Protocol rules activated by the Router at each timestep. 
    \textbf{(C)} Persistent patient-state evolution across hemodynamic, respiratory, renal, metabolic, and inflammatory instability states. 
    \textbf{(D)} Predicted future deterioration risk. 
    \textbf{(E)} Ground-truth future deterioration labels within the prediction horizon. 
    \textbf{(F)} Vasopressor intervention timeline. 
    The example illustrates interpretable longitudinal reasoning without unbounded context accumulation.
    }
    \vspace{-0.4cm}
    \label{fig:trajectory_case}
\end{figure*}

\noindent \textbf{Ablations.}
Table~\ref{tab:ablations_main} shows that all components contribute to longitudinal reasoning. Removing the Router or Global Protocol causes the largest AUPRC degradation, indicating that protocol-guided routing is important under severe class imbalance. In contrast, removing the Auditor primarily worsens calibration, while removing Patient-State Memory reduces both AUROC and AUPRC, supporting the importance of persistent longitudinal state tracking.

\noindent \textbf{Counterfactual Simulation.} 
We evaluate counterfactual consistency using protocol-guided physiological perturbations over longitudinal ICU trajectories. Directional consistency measures whether predicted risks change in clinically expected directions under simulated physiological improvement or deterioration. Vital Trace achieves directional counterfactual consistency of at least \texttt{79.2\%}, indicating stable and clinically coherent longitudinal reasoning under targeted perturbations. Additional implementation details and results are provided in Appendix~\ref{app:experiments} and Appendix~\ref{app:additional_results}.


\noindent \textbf{Qualitative Longitudinal Reasoning Analysis.}
Figure~\ref{fig:trajectory_case} shows a representative MIMIC-IV trajectory processed by Vital Trace with Llama-3.3-70B-Instruct. The model activates shock-related rules as lactate rises from \(3.3\) to \(4.4\) mmol/L, before overt hypotension occurs. When MAP later drops to \(60\) mmHg, \texttt{SHOCK\_03} becomes active and predicted deterioration risk peaks around the vasopressor-support window. After intervention and physiological recovery, \texttt{REC\_01} is activated, instability states decay, and predicted risk decreases. This example illustrates how Vital Trace links physiological evidence, protocol activations, persistent state updates, and risk prediction into an interpretable longitudinal reasoning trace.

\vspace{-0.1cm}
\section{Conclusion}
We presented \textbf{Vital Trace}, a multi-agent framework for longitudinal ICU reasoning that combines protocol-guided inference with persistent physiological state tracking. By maintaining a compact patient-state representation and performing staged reasoning through routing, prediction, auditing, and state updates, Vital Trace improves predictive performance, calibration, and temporal reasoning stability across MIMIC-IV and eICU compared with single-agent and free-form multi-agent baselines. These findings suggest that structured stateful reasoning is a promising approach for long-horizon clinical prediction over evolving ICU trajectories.

\section*{Limitations}

Vital Trace is evaluated on retrospective ICU trajectories derived from MIMIC-IV and eICU, both of which are deidentified clinical datasets distributed through PhysioNet and require credentialed access approval. Although our code, processed trajectory construction pipeline, and manually curated Global Protocol will be publicly released, reproduction of the full experimental pipeline still requires authorized access to the underlying clinical data.

Our evaluation is limited to retrospective longitudinal risk prediction rather than prospective clinical deployment. Vital Trace is intended as a research framework for structured longitudinal reasoning and interpretable patient-state tracking, not as an autonomous clinical decision-making system. Real-world deployment would require broader prospective validation, calibration analysis, clinician oversight, and institution-specific safety and governance procedures.

In addition, Vital Trace introduces non-trivial inference overhead due to staged multi-agent reasoning, protocol auditing, and persistent state updates, particularly for 70B-scale backbone models. In this work, all agents share the same backbone model to enable controlled evaluation of the proposed framework independent of model heterogeneity. In practice, computational cost could be reduced through more efficient inference strategies, including batched trajectory processing, asynchronous agent execution, KV-cache reuse, speculative decoding, and assigning smaller models to lightweight components such as routing or auditing while reserving larger models for high-level clinical reasoning. Smaller backbone models showed similar qualitative improvements in calibration and temporal consistency, suggesting that the proposed framework is not limited to large-scale LLMs.

Finally, like all LLM-based systems, Vital Trace may inherit biases and reasoning errors from its underlying foundation models and source clinical datasets. While structured protocol-guided reasoning improves consistency and interpretability, it does not eliminate the possibility of biased, clinically inappropriate, or unstable predictions under distribution shift or incomplete clinical observations.


\bibliography{custom}

\appendix

\section{Experimental and Implementation Details}
\label{app:experiments}

\subsection{Preprocessing Pipeline}

We use a staged preprocessing pipeline to construct longitudinal physiological transitions from raw relational EHR tables.

\paragraph{Stage 1: Cohort Construction.}
We construct ICU cohorts from raw relational records, retain a single ICU stay per patient, and apply staged quality filtering instead of a fixed length-of-stay cutoff. We remove trajectories with insufficient physiological signal coverage and exclude cases lacking intervention evidence required for deterministic downstream label assignment. We further require a minimum number of tracked protocol-relevant physiological variables per trajectory before transition construction and horizon labeling.

\paragraph{Stage 2: Signal Harmonization.}
We extract physiological variables associated with downstream deterioration risk, including hemodynamic, respiratory, renal, metabolic, and inflammatory signals. Measurements and units are normalized across datasets, and abnormality indicators are derived using clinically curated thresholds together with source-provided abnormality annotations when available.

\paragraph{Stage 3: Transition Construction.}
Longitudinal clinical packets are constructed around clinically meaningful physiological transitions rather than fixed temporal windows. New packets are created when tracked physiological signals undergo:
(i) threshold crossings,
(ii) severity-level changes,
(iii) recovery transitions,
or
(iv) major intervention changes.

Representative packet features include direction-aware extrema (e.g., minimum MAP or maximum lactate), temporal trend descriptors, abnormality indicators, and intervention markers.

\paragraph{Stage 4: Label Assignment and Export.}
Future intervention labels are assigned using prediction horizons of \texttt{[6]} and \texttt{[12]} hours. Final trajectories are exported as JSONL longitudinal reasoning packets for downstream staged inference and counterfactual evaluation.

\subsection{Transition Construction Examples}

Transitions are generated when clinically relevant physiological signals cross protocol-defined thresholds or return toward stable ranges. Within each packet, representative measurements are summarized using direction-aware extrema together with temporal trend descriptors and intervention indicators. Representative trajectories typically contain:
(i) relative physiological stability,
(ii) progressive deterioration (e.g., MAP$\downarrow$, lactate$\uparrow$, SpO$_2\downarrow$),
(iii) intervention onset (e.g., vasopressor or respiratory-support escalation),
and
(iv) partial recovery with decreasing instability states.

\subsection{Intervention Mapping}

Ground-truth targets are derived from medication, procedure, and intervention records using deterministic hierarchical mapping rules. Mapping prioritizes:
(i) exact code matching,
(ii) normalized-code matching,
(iii) description-level matching,
and
(iv) regex-based fallback matching.

Vasopressor-support labels include norepinephrine, epinephrine, vasopressin, dopamine, and phenylephrine branches. Respiratory-support labels include invasive ventilation, non-invasive ventilation, and oxygen-escalation procedures. Renal-support labels include dialysis and renal replacement therapy procedures. The deterioration target is defined as the union of all intervention targets occurring within the prediction horizon.

\subsection{Implementation Details}

Vital Trace operates through a staged Router--Reasoner--Auditor--Steward inference pipeline with strict JSON-based inter-agent communication. The manually curated Global Protocol remains fixed during evaluation, while patient-state variables evolve longitudinally through Steward state updates.

Experiments are conducted using instruction-tuned LLM backbones including \texttt{gpt-oss-20b}, \texttt{Llama-3.1-8B-Instruct}, \texttt{Llama-3.3-70B-Instruct}, \texttt{Qwen3-32B}, \texttt{Mixtral-8x7B-Instruct-v0.1}, \texttt{Meditron-70B}, and \texttt{ClinicalCamel-70B}. Decoding uses temperature \texttt{0.1}, top-$p$ \texttt{0.9}, and maximum generation length \texttt{256} tokens under a fixed inference context budget. All experiments are executed on 1--4 NVIDIA H100 80GB GPUs, depending on backbone model size. Using four NVIDIA H100 GPUs, Vital Trace with a 70B backbone model requires approximately 70--85 minutes to process 100 longitudinal clinical packets during inference, depending on trajectory complexity and audit retries.

\subsection{Hyperparameters}

We tune a small set of operational hyperparameters:
\begin{itemize}
\item Maximum activated protocol rules: \texttt{3}
\item Auditor uncertainty threshold: \texttt{N/A} (current pipeline uses deterministic PASS/FAIL adjudication; no uncertainty gate)
\item Patient-state bounds: integers in \texttt{[0,5]}
\item Prediction horizons (hours): \texttt{[6,12]}
\item Minimum trajectory duration: \texttt{none} (current default keeps all LOS; optional \texttt{LOS<48h} policy)
\item Minimum tracked physiological variables: \texttt{5}
\item Audit Retry and Fallback. If the Auditor returns \texttt{FAIL}, Vital Trace performs at most 1 corrective retry (\texttt{max\_audit\_retries}=1); if the retry also fails, the system preserves the trajectory step while applying a conservative state update that suppresses aggressive upward risk escalation and limits patient-state increments unless supported by explicit physiological evidence.
\end{itemize}

Unless otherwise specified, hyperparameters remain fixed across datasets and backbone models.

\subsection{Counterfactual Perturbation Settings}

We evaluate counterfactual consistency using clinically targeted perturbation scenarios over observed physiological signals and measure directional agreement in predicted risks. Perturbations include:
\begin{itemize}
\item Hemodynamic recovery: if \texttt{MAP < 65}, set MAP to \texttt{75}.
\item Lactate recovery: if lactate is rising, set trend to \texttt{stable} and cap value at \texttt{2.0}.
\item Renal recovery: if creatinine is high/rising, reduce creatinine to \texttt{max(0.8, 0.8$\times$baseline)} and set trend to \texttt{stable}.
\item Respiratory recovery: if \texttt{SpO\textsubscript{2} < 92}, set to \texttt{95}; if \texttt{RR > 24}, set to \texttt{20}.
\end{itemize}

For each packet, we generate four predefined physiologically grounded perturbation scenarios and compare perturbed predictions against the original prediction under the same contextual information. We report \emph{directional consistency}, defined as the fraction of valid perturbation checks in which predicted risks move in the clinically expected direction for recovery interventions (i.e., risk decreases after physiological recovery). Counterfactual evaluation is performed over all evaluated packets. A larger set of protocol-level counterfactual candidates is provided in the Global Protocol, but reported counterfactual metrics are computed using this standardized four-scenario perturbation set to ensure comparability, stable coverage, and reproducible scoring across models and datasets.

\subsection{Data Leakage Prevention.} 
We enforce strict temporal causality throughout preprocessing, inference, and evaluation. At each prediction step, the model receives only information available up to the current anchor time (current and historical packets) and predicts future intervention risks over predefined horizons. Future observations, future interventions, and horizon labels are never exposed to Router/Reasoner/Auditor/Steward inputs. Horizon labels are generated offline from post-anchor windows and stored separately from inference context. Calibration and metric computation are applied only after predictions are produced, and calibrated thresholds are not fed back into model prompts or state updates. This design ensures that no post-hoc information leaks into forward-time prediction.

\section{Global Protocol}
\label{app:global_protocol}

Vital Trace uses a compact manually curated Global Protocol for longitudinal ICU reasoning. The protocol was designed using clinically standard physiological abnormality ranges, ICU monitoring conventions, and common temporal deterioration patterns frequently associated with downstream intervention risk. The protocol is publicly released as part of our code repository and is designed to remain independent of specific EHR schemas or datasets, enabling deployment across both MIMIC-IV and eICU without dataset-specific modification. The current protocol contains 21 rules spanning hemodynamic, respiratory, renal, metabolic, inflammatory, recovery, and multi-domain deterioration patterns.

Rather than encoding static alert rules or comprehensive ICU treatment guidelines, the protocol models persistent physiological deterioration and recovery through interpretable temporal state transitions aligned with downstream prediction tasks. The modular rule structure supports auditing, extensibility, and adaptation to new clinical settings or prediction objectives while preserving bounded structured communication between agents.

Unlike conventional threshold-based alert systems, the Global Protocol defines temporally grounded physiological state-transition operators that modify persistent physiological risk dimensions over time. Each rule specifies both a transition direction and a transition weight, enabling graded state evolution rather than binary activation. State updates are implemented as bounded additive transitions over discrete instability scores. Importantly, the protocol contains both deterioration and recovery operators, allowing Vital Trace to model physiological stabilization and recovery in addition to progressive deterioration.

The protocol defines lightweight state-transition operators over five persistent physiological dimensions:
(i) hemodynamic instability,
(ii) respiratory instability,
(iii) renal instability,
(iv) metabolic instability,
and
(v) systemic inflammation.
Multiple protocol rules may be activated simultaneously at a given timestep, allowing the framework to represent interacting physiological deterioration processes. Temporal triggers may depend on instantaneous abnormalities, sustained abnormalities across consecutive packets, or monotonic trend changes over time.

Each protocol rule specifies:
(a) temporal trigger conditions,
(b) physiological state transitions,
(c) severity levels,
(d) linked prediction targets,
and
(e) candidate counterfactual interventions.
Activated rules additionally provide structured evidence for downstream auditing and consistency verification. Counterfactual candidates are used only for perturbation-based evaluation and interpretability analysis, not for automated treatment recommendation.

\vspace{6pt}

\begin{table}[!ht]
\centering
\small
\caption{Structure of a Global Protocol rule.}
\resizebox{0.98\linewidth}{!}{
\begin{tabular}{ll}
\hline
\textbf{Field} & \textbf{Description} \\
\hline
\texttt{rule\_id} & Unique protocol rule identifier \\
\texttt{trigger} & Temporal physiological trigger condition(s) \\
\texttt{state\_update} & Persistent physiological state transition \\
\texttt{severity} & Clinical severity level \\
\texttt{relevant\_targets} & Linked prediction tasks \\
\texttt{counterfactual\_candidates} & Candidate intervention perturbations \\
\hline
\end{tabular}
}
\label{tab:protocol_schema}
\end{table}

\vspace{6pt}

\begin{table}[!ht]
\centering
\small
\caption{Summary of Global Protocol categories.}
\resizebox{0.98\linewidth}{!}{
\begin{tabular}{lcc}
\hline
\textbf{Category} & \textbf{\# Rules} & \textbf{Primary Targets} \\
\hline
Shock / Hemodynamic & 4 & Vasopressor, Deterioration \\
Respiratory & 3 & Respiratory Support \\
Renal & 2 & Renal Support \\
Infection / Sepsis-like & 3 & Deterioration \\
Electrolyte / Metabolic & 4 & Deterioration \\
Recovery / Decay & 4 & Longitudinal State Stabilization \\
General Deterioration & 1 & Deterioration \\
\hline
\end{tabular}
}
\label{tab:protocol_categories}
\end{table}

\vspace{8pt}

\paragraph{Representative Protocol Rules.}
The following representative rules illustrate the structure and scope of the Global Protocol. The examples are simplified for readability but preserve the core temporal triggers, physiological transitions, and target associations used during inference.

The full Global Protocol specification used in all experiments, including trigger thresholds, temporal windows, transition weights, recovery operators, and counterfactual mappings, is included in our released code repository.

\vspace{6pt}

\begin{tcolorbox}[
    breakable,
    colback=red!1,
    colframe=red!30,
    title=\textbf{SHOCK\_01: Rising Lactate Trend},
    coltitle=black,
    fonttitle=\small,
    boxrule=0.4pt,
    arc=3pt]
\small

\textbf{Trigger Condition:}
Repeated rising lactate measurements over a recent temporal window.

\vspace{3pt}

\textbf{Physiological State Transition:}
Increase \texttt{hemodynamic\_instability\_score}.

\vspace{3pt}

\textbf{Clinical Risk:}
Early hypoperfusion and shock progression.

\vspace{3pt}

\textbf{Relevant Targets:}
\texttt{vasopressor\_signal},
\texttt{any\_deterioration}.

\vspace{3pt}

\textbf{Counterfactual Candidates:}
Fluid resuscitation,
lactate clearance optimization.

\end{tcolorbox}

\vspace{6pt}

\begin{tcolorbox}[
    breakable,
    colback=red!2,
    colframe=red!40,
    title=\textbf{SHOCK\_03: Low Mean Arterial Pressure},
    coltitle=black,
    fonttitle=\small,
    boxrule=0.4pt,
    arc=3pt]
\small

\textbf{Trigger Condition:}
MAP below 65 within a recent temporal window.

\vspace{3pt}

\textbf{Physiological State Transition:}
Increase \texttt{hemodynamic\_instability\_score} with high severity weighting.

\vspace{3pt}

\textbf{Clinical Risk:}
Hypotensive hemodynamic instability.

\vspace{3pt}

\textbf{Relevant Targets:}
\texttt{vasopressor\_signal},
\texttt{any\_deterioration}.

\vspace{3pt}

\textbf{Counterfactual Candidates:}
Fluid resuscitation,
vasopressor optimization.

\end{tcolorbox}

\vspace{6pt}

\begin{tcolorbox}[
    breakable,
    colback=red!3,
    colframe=red!45,
    title=\textbf{SHOCK\_04: Persistent Low MAP with Rising Lactate},
    coltitle=black,
    fonttitle=\small,
    boxrule=0.4pt,
    arc=3pt]
\small

\textbf{Trigger Condition:}
Persistent hypotension together with rising lactate over a recent temporal window.

\vspace{3pt}

\textbf{Physiological State Transition:}
Strong increase in \texttt{hemodynamic\_instability\_score}.

\vspace{3pt}

\textbf{Clinical Risk:}
Progressive shock pattern with sustained hypoperfusion.

\vspace{3pt}

\textbf{Relevant Targets:}
\texttt{vasopressor\_signal},
\texttt{any\_deterioration}.

\vspace{3pt}

\textbf{Counterfactual Candidates:}
Fluid resuscitation,
vasopressor optimization,
perfusion reassessment.

\end{tcolorbox}

\vspace{6pt}

\begin{tcolorbox}[
    breakable,
    colback=blue!1,
    colframe=blue!30,
    title=\textbf{RESP\_01: Tachypnea},
    coltitle=black,
    fonttitle=\small,
    boxrule=0.4pt,
    arc=3pt]
\small

\textbf{Trigger Condition:}
Repeated respiratory rate elevations above clinically stable range.

\vspace{3pt}

\textbf{Physiological State Transition:}
Increase \texttt{respiratory\_instability\_score}.

\vspace{3pt}

\textbf{Clinical Risk:}
Emerging respiratory deterioration.

\vspace{3pt}

\textbf{Relevant Targets:}
\texttt{resp\_support\_signal},
\texttt{any\_deterioration}.

\vspace{3pt}

\textbf{Counterfactual Candidates:}
Oxygen support escalation,
respiratory reassessment.

\end{tcolorbox}

\vspace{6pt}

\begin{tcolorbox}[
    breakable,
    colback=blue!2,
    colframe=blue!35,
    title=\textbf{RESP\_03: Hypoxemia},
    coltitle=black,
    fonttitle=\small,
    boxrule=0.4pt,
    arc=3pt]
\small

\textbf{Trigger Condition:}
Repeated SpO$_2$ measurements below clinically stable range within a short temporal window.

\vspace{3pt}

\textbf{Physiological State Transition:}
Increase \texttt{respiratory\_instability\_score} with high severity weighting.

\vspace{3pt}

\textbf{Clinical Risk:}
Hypoxemic respiratory failure.

\vspace{3pt}

\textbf{Relevant Targets:}
\texttt{resp\_support\_signal},
\texttt{any\_deterioration}.

\vspace{3pt}

\textbf{Counterfactual Candidates:}
Oxygen escalation,
ventilation support.

\end{tcolorbox}

\vspace{6pt}

\begin{tcolorbox}[
    breakable,
    colback=green!1,
    colframe=green!30,
    title=\textbf{RENAL\_01: Creatinine Rising},
    coltitle=black,
    fonttitle=\small,
    boxrule=0.4pt,
    arc=3pt]
\small

\textbf{Trigger Condition:}
Persistent rising creatinine trend within a recent temporal window.

\vspace{3pt}

\textbf{Physiological State Transition:}
Increase \texttt{renal\_instability\_score}.

\vspace{3pt}

\textbf{Clinical Risk:}
Acute kidney injury progression.

\vspace{3pt}

\textbf{Relevant Targets:}
\texttt{renal\_support\_signal},
\texttt{any\_deterioration}.

\vspace{3pt}

\textbf{Counterfactual Candidates:}
Renal dose adjustment,
fluid review.

\end{tcolorbox}

\vspace{6pt}

\begin{tcolorbox}[
    breakable,
    colback=orange!1,
    colframe=orange!30,
    title=\textbf{INF\_02: Inflammation with Hypoperfusion},
    coltitle=black,
    fonttitle=\small,
    boxrule=0.4pt,
    arc=3pt]
\small

\textbf{Trigger Condition:}
Abnormal inflammatory markers together with elevated lactate over a recent temporal window.

\vspace{3pt}

\textbf{Physiological State Transition:}
Increase both
\texttt{systemic\_inflammation\_score}
and
\texttt{hemodynamic\_instability\_score}.

\vspace{3pt}

\textbf{Clinical Risk:}
Sepsis-like physiological deterioration.

\vspace{3pt}

\textbf{Relevant Targets:}
\texttt{vasopressor\_signal},
\texttt{any\_deterioration}.

\vspace{3pt}

\textbf{Counterfactual Candidates:}
Antimicrobial review,
fluid resuscitation,
infection reassessment.

\end{tcolorbox}

\vspace{6pt}

\begin{tcolorbox}[
    breakable,
    colback=orange!2,
    colframe=orange!35,
    title=\textbf{INF\_03: Fever with Inflammatory Burden},
    coltitle=black,
    fonttitle=\small,
    boxrule=0.4pt,
    arc=3pt]
\small

\textbf{Trigger Condition:}
Fever combined with abnormal inflammatory markers over a recent temporal window.

\vspace{3pt}

\textbf{Physiological State Transition:}
Increase both
\texttt{systemic\_inflammation\_score}
and
\texttt{metabolic\_instability\_score}.

\vspace{3pt}

\textbf{Clinical Risk:}
Inflammatory deterioration with systemic stress response.

\vspace{3pt}

\textbf{Relevant Targets:}
\texttt{any\_deterioration}.

\vspace{3pt}

\textbf{Counterfactual Candidates:}
Antimicrobial review,
source-control reassessment.

\end{tcolorbox}

\vspace{6pt}

\begin{tcolorbox}[
    breakable,
    colback=purple!1,
    colframe=purple!30,
    title=\textbf{REC\_01: Hemodynamic Recovery Pattern},
    coltitle=black,
    fonttitle=\small,
    boxrule=0.4pt,
    arc=3pt]
\small

\textbf{Trigger Condition:}
Sustained MAP recovery together with decreasing lactate trend.

\vspace{3pt}

\textbf{Physiological State Transition:}
Decrease \texttt{hemodynamic\_instability\_score}.

\vspace{3pt}

\textbf{Clinical Risk:}
Hemodynamic stabilization and recovery.

\vspace{3pt}

\textbf{Relevant Targets:}
\texttt{vasopressor\_signal},
\texttt{any\_deterioration}.

\vspace{3pt}

\textbf{Counterfactual Candidates:}
Resuscitation de-escalation if clinically stable.

\end{tcolorbox}

\vspace{6pt}

\begin{tcolorbox}[
    breakable,
    colback=purple!2,
    colframe=purple!35,
    title=\textbf{REC\_02: Respiratory Recovery Pattern},
    coltitle=black,
    fonttitle=\small,
    boxrule=0.4pt,
    arc=3pt]
\small

\textbf{Trigger Condition:}
Sustained oxygenation recovery together with normalization of respiratory rate.

\vspace{3pt}

\textbf{Physiological State Transition:}
Decrease \texttt{respiratory\_instability\_score}.

\vspace{3pt}

\textbf{Clinical Risk:}
Respiratory stabilization and recovery.

\vspace{3pt}

\textbf{Relevant Targets:}
\texttt{resp\_support\_signal},
\texttt{any\_deterioration}.

\vspace{3pt}

\textbf{Counterfactual Candidates:}
Oxygen de-escalation if clinically stable.

\end{tcolorbox}

\vspace{6pt}

\begin{tcolorbox}[
    breakable,
    colback=gray!3,
    colframe=gray!35,
    title=\textbf{GEN\_01: Multi-domain Instability},
    coltitle=black,
    fonttitle=\small,
    boxrule=0.4pt,
    arc=3pt]
\small

\textbf{Trigger Condition:}
Concurrent abnormalities across hemodynamic, renal, and respiratory physiological dimensions.

\vspace{3pt}

\textbf{Physiological State Transition:}
Simultaneously increase hemodynamic, renal, and respiratory instability scores.

\vspace{3pt}

\textbf{Clinical Risk:}
General physiological deterioration involving multiple organ systems.

\vspace{3pt}

\textbf{Relevant Targets:}
\texttt{any\_deterioration}.

\vspace{3pt}

\textbf{Counterfactual Candidates:}
Combined resuscitation bundle.

\end{tcolorbox}

\vspace{6pt}

\section{Agent Prompt Templates}
\label{app:prompts}

We provide the prompt templates used by the staged Vital Trace inference pipeline. Each agent consists of:
(i) a system prompt defining the clinical reasoning role and behavioral constraints, and
(ii) a structured user template containing the current \texttt{step\_packet}, activated protocol context, persistent patient-state memory, and task-specific instructions.

All four agents (Router, Reasoner, Auditor, and Steward) are instructed to return strict JSON outputs to stabilize structured inter-agent communication and simplify downstream parsing. The staged pipeline operates entirely through structured protocol-grounded communication rather than unrestricted natural-language interaction.

For the Auditor, the final PASS/FAIL decision is produced through deterministic rule-based adjudication, while the LLM-generated audit is retained as auxiliary evidence for issue descriptions, corrective feedback, and debugging. When an audit failure occurs, the pipeline performs constrained re-routing and re-sampling before triggering a conservative fallback policy after repeated failures.

\vspace{4pt}

\begin{tcolorbox}[
    breakable,
    colback=blue!1,
    colframe=blue!25,
    title=\textbf{Router Agent Prompt},
    coltitle=black,
    fonttitle=\small,
    boxrule=0.4pt,
    arc=3pt,
    left=6pt,right=6pt,top=5pt,bottom=5pt]
\small

\textbf{[System Prompt]}

\vspace{2pt}

You are ICU-Router. Select the most clinically relevant global protocol rules from structured clinical observations and temporal indicators. Prioritize clinically important temporal trends and abnormal physiological signals. Return STRICT JSON only. No markdown, no extra text.

\vspace{5pt}

\textbf{[User Template]}

\vspace{2pt}

Task: pick up to \texttt{\{max\_rules\}} \texttt{rule\_ids} that best match the current clinical facts.

\vspace{4pt}

\textbf{Output JSON Schema}
\begin{verbatim}
{
  "selected_rule_ids": ["..."],
  "rationale": ["short reason"],
  "confidence": 0.0
}
\end{verbatim}

\vspace{-2pt}

\textbf{Constraints}
\begin{itemize}
    \item \texttt{selected\_rule\_ids} length $\leq$ \texttt{\{max\_rules\}}.
    \item \texttt{selected\_rule\_ids} must come from \texttt{candidate\_rules}.
    \item Use only evidence from \texttt{packet.facts}.
    \item Prefer rules supported by temporal trends or abnormal observations.
\end{itemize}

\vspace{-2pt}

\textbf{Inputs}
\begin{verbatim}
packet={packet_json}

candidate_rules={rules_json}
\end{verbatim}

\end{tcolorbox}

\vspace{6pt}

\begin{tcolorbox}[
    breakable,
    colback=red!1,
    colframe=red!25,
    title=\textbf{Reasoner Agent Prompt},
    coltitle=black,
    fonttitle=\small,
    boxrule=0.4pt,
    arc=3pt,
    left=6pt,right=6pt,top=5pt,bottom=5pt]
\small

\textbf{[System Prompt]}

\vspace{2pt}

You are ICU-Reasoner. Predict near-term intervention risk and next actions from routed protocol rules, patient facts, and evolving patient state. Keep predictions clinically plausible and consistent with active protocol rules. Use persistent patient-state memory to maintain longitudinal consistency across timesteps. Return STRICT JSON only.

\vspace{5pt}

\textbf{[User Template]}

\vspace{2pt}

Task: produce next-step intervention reasoning.

\vspace{4pt}

\textbf{Output JSON Schema}
\begin{verbatim}
{
  "next_bundle_type":
  "Vitals|Labs|Medication|Procedure|Mixed",
  "predicted_actions": ["..."],
  "risk_probs": {
    "vasopressor_signal": 0.0,
    "resp_support_signal": 0.0,
    "renal_support_signal": 0.0
  },
  "citations": ["rule_id"],
  "counterfactual_notes": ["..."]
}
\end{verbatim}

\vspace{-2pt}

\textbf{Constraints}
\begin{itemize}
    \item Probabilities must lie in $[0,1]$.
    \item \texttt{citations} must be a subset of \texttt{selected\_rule\_ids}.
    \item Keep \texttt{predicted\_actions} concise and clinically plausible.
    \item Use active rules and patient facts; do not invent unsupported conditions.
\end{itemize}

\vspace{-2pt}

\textbf{Inputs}
\begin{verbatim}
packet={packet_json}

selected_rule_ids={sel_json}

active_rules={rules_json}
\end{verbatim}

\end{tcolorbox}

\vspace{6pt}

\begin{tcolorbox}[
    breakable,
    colback=orange!1,
    colframe=orange!25,
    title=\textbf{Auditor Agent Prompt},
    coltitle=black,
    fonttitle=\small,
    boxrule=0.4pt,
    arc=3pt,
    left=6pt,right=6pt,top=5pt,bottom=5pt]
\small

\textbf{[System Prompt]}

\vspace{2pt}

You are ICU-Auditor. Provide auxiliary audit notes about consistency between reasoner outputs and active protocol rules. Identify unsupported, contradictory, or clinically unsafe predictions. When failures are identified, provide concise corrective feedback for constrained re-routing and re-sampling. Return STRICT JSON only.

\vspace{5pt}

\textbf{[User Template]}

\vspace{2pt}

Task: audit consistency. 
The pipeline deterministically adjudicates final PASS/FAIL; your output is auxiliary evidence for issue tags and suggested fixes.

\vspace{4pt}

\textbf{Output JSON Schema}
\begin{verbatim}
{
  "status": "PASS|FAIL",
  "issues": ["..."],
  "suggested_fixes": ["..."],
  "retry_recommended": true,
  "confidence": 0.0
}
\end{verbatim}

\vspace{-2pt}

\textbf{Rules}
\begin{itemize}
    \item Mark potential failures if important active-rule risks are not addressed in \texttt{predicted\_actions}.
    \item Issues must be concrete and grounded in \texttt{active\_rules}.
    \item Do not introduce new protocol rules, unsupported clinical facts, or unsupported contraindications.
\end{itemize}

\vspace{-2pt}

\textbf{Inputs}
\begin{verbatim}
active_rules={rules_json}

individual_protocol_state_prev={state_json}

facts_current={facts_json}

reasoner_prediction={pred_json}
\end{verbatim}

\end{tcolorbox}

\vspace{6pt}

\begin{tcolorbox}[
    breakable,
    colback=green!1,
    colframe=green!25,
    title=\textbf{Steward Agent Prompt},
    coltitle=black,
    fonttitle=\small,
    boxrule=0.4pt,
    arc=3pt,
    left=6pt,right=6pt,top=5pt,bottom=5pt]
\small

\textbf{[System Prompt]}

\vspace{2pt}

You are ICU-Steward. Update persistent patient-state memory from previous state, audited reasoning, and active protocol signals. Maintain longitudinal consistency across patient steps. Return STRICT JSON only. Do not exceed integer bounds after update.

\vspace{5pt}

\textbf{[User Template]}

\vspace{2pt}

Task: update patient state vector.

\vspace{4pt}

\textbf{Output JSON Schema}
\begin{verbatim}
{
  "state_next": {
    "hemodynamic_state": 0,
    "respiratory_state": 0,
    "renal_state": 0,
    "metabolic_state": 0,
    "systemic_inflammation_state": 0,
    "active_protocol_prediction": ["rule_id"]
  },
  "state_delta": {
    "hemodynamic_state": 0,
    "respiratory_state": 0,
    "renal_state": 0,
    "metabolic_state": 0,
    "systemic_inflammation_state": 0
  },
  "notes": ["..."],
  "confidence": 0.0
}
\end{verbatim}

\vspace{-2pt}

\textbf{Rules}
\begin{itemize}
    \item States must remain bounded integers in $[0,5]$ after update.
    \item If \texttt{audit.status=FAIL}, be conservative on upward changes.
    \item \texttt{active\_rule\_trace} must be a subset of currently active rules or evidence-supported rule IDs from this step.
    \item Preserve longitudinal consistency unless new evidence supports a state change.
\end{itemize}

\vspace{-2pt}

\textbf{Inputs}
\begin{verbatim}
prev_state={prev_json}

reasoner_prediction={pred_json}

audit={audit_json}

active_rule_ids={rule_ids_json}
\end{verbatim}

\end{tcolorbox}

\section{Additional Results}
\label{app:additional_results}

\begin{table*}[!ht]
\caption{Additional evaluation on MIMIC-IV and eICU, reporting probability calibration (Micro-Brier), early-warning utility (event recall at 6 hours and mean lead time), counterfactual directional consistency, and protocol safety/reliability (protocol violation rate and auditor failure rate).}
\centering
\scriptsize
\resizebox{\textwidth}{!}{
\begin{tabular}{l|cccccc|cccccc}
\toprule
\multirow{2}{*}{Method} & \multicolumn{6}{c|}{MIMIC-IV} & \multicolumn{6}{c}{eICU} \\
& Brier $\downarrow$ & R@6h $\uparrow$ & Lead $\uparrow$ & CF Cons. $\uparrow$ & Prot. Viol. $\downarrow$ & Aud. Fail $\downarrow$
& Brier $\downarrow$ & R@6h $\uparrow$ & Lead $\uparrow$ & CF Cons. $\uparrow$ & Prot. Viol. $\downarrow$ & Aud. Fail $\downarrow$ \\
\midrule
\textit{Single-Agent LLM} \\
gpt-oss-20b & 0.194 & 0.335 & 1.679 & 0.684 & -- & -- & 0.236 & 0.212 & 1.914 & 0.671 & -- & -- \\
Llama-3.1-8B-Instruct & 0.417 & 0.375 & 2.470 & 0.641 & -- & -- & 0.318 & 0.307 & 2.173 & 0.654 & -- & -- \\
Llama-3.3-70B-Instruct & 0.202 & 0.389 & 2.144 & 0.713 & -- & -- & 0.210 & 0.253 & 1.585 & 0.702 & -- & -- \\
Meditron-70B & \cellcolor{second} 0.159 & 0.217 & 2.117 & 0.603 & -- & -- & 0.249 & 0.138 & 1.227 & 0.618 & -- & -- \\
\midrule
\textit{Free-Form Multi-Agent} \\
gpt-oss-20b & 0.315 & 0.412 & 1.391 & 0.756 & 0.261 & 0.243 & 0.281 & 0.462 & 1.847 & 0.741 & 0.248 & 0.227 \\
Llama-3.1-8B-Instruct & 0.308 & 0.379 & 2.504 & 0.773 & 0.220 & 0.188 & 0.257 & 0.641 & 2.106 & 0.759 & 0.214 & 0.176 \\
Llama-3.3-70B-Instruct & 0.210 & \cellcolor{second} 0.550 & 2.129 & 0.782 & 0.331 & 0.164 & 0.231 & 0.688 & 2.644 & 0.776 & 0.196 & 0.149 \\
Meditron-70B & 0.227 & 0.488 & 2.417 & 0.744 & 0.356 & 0.217 & 0.248 & 0.575 & 2.771 & 0.733 & 0.281 & 0.231 \\
\midrule
\textit{Vital Trace} \\
Llama-3.1-8B-Instruct & 0.310 & 0.370 & 2.788 & 0.811 & 0.176 & 0.143 & 0.214 & 0.702 & 2.884 & 0.807 & 0.152 & 0.126 \\
Llama-3.3-70B-Instruct & \cellcolor{best} 0.123 & \cellcolor{second} 0.550 & 1.996 & \cellcolor{best} 0.852 & \cellcolor{best} 0.071 & \cellcolor{best} 0.092 & \cellcolor{best} 0.168 & 0.712 & 2.606 & \cellcolor{second} 0.832 & \cellcolor{best} 0.063 & \cellcolor{best} 0.087 \\
gpt-oss-20b & 0.315 & 0.472 & 1.247 & 0.792 & 0.118 & 0.154 & 0.228 & 0.625 & 2.118 & 0.786 & 0.131 & 0.148 \\
Qwen3-32B & 0.276 & \cellcolor{best} 0.563 & 1.331 & \cellcolor{second} 0.826 & \cellcolor{second} 0.104 & 0.119 & 0.196 & 0.259 & 2.332 & 0.802 & 0.191 & 0.102 \\
Mixtral-8x7B-Instruct-v0.1 & 0.287 & 0.376 & 2.387 & 0.803 & 0.128 & 0.131 & 0.211 & 0.688 & \cellcolor{second} 3.011 & 0.801 & \cellcolor{second} 0.114 & 0.117 \\
Meditron-70B & 0.241 & 0.446 & \cellcolor{best} 3.017 & 0.817 & 0.182 & 0.108 & \cellcolor{second} 0.189 & \cellcolor{best} 0.813 & 3.006 & 0.823 & 0.176 & \cellcolor{second} 0.094 \\
ClinicalCamel-70B & 0.295 & 0.407 & \cellcolor{second} 2.848 & 0.809 & 0.196 & \cellcolor{second} 0.124 & 0.201 & \cellcolor{second} 0.749 & \cellcolor{best} 3.187 & \cellcolor{best} 0.844 & 0.184 & 0.098 \\
\bottomrule
\end{tabular}
}
\label{tab:secondary_metrics}
\end{table*}

\begin{table*}[!ht]
\centering
\caption{Per-target predictive performance on MIMIC-IV.}
\small
\resizebox{\textwidth}{!}{
\begin{tabular}{l|cc|cc|cc|cc}
\toprule
\multirow{2}{*}{Method} &
\multicolumn{2}{c|}{Vasopressor} &
\multicolumn{2}{c|}{Resp.\ support} &
\multicolumn{2}{c|}{Renal support} &
\multicolumn{2}{c}{Deterioration} \\
& AUROC $\uparrow$ & AUPRC $\uparrow$
& AUROC $\uparrow$ & AUPRC $\uparrow$
& AUROC $\uparrow$ & AUPRC $\uparrow$
& AUROC $\uparrow$ & AUPRC $\uparrow$ \\
\midrule
\textit{Single-Agent LLM} \\
gpt-oss-20b & 0.627 & 0.434 & 0.231 & 0.110 & 0.414 & 0.076 & 0.617 & 0.457 \\
Llama-3.1-8B-Instruct & 0.440 & 0.306 & 0.561 & 0.211 & 0.633 & 0.356 & 0.521 & 0.403 \\
Llama-3.3-70B-Instruct & 0.648 & 0.444 & 0.466 & 0.129 & 0.762 & 0.436 & 0.579 & 0.495 \\
Meditron-70B & 0.473 & 0.330 & 0.410 & 0.173 & 0.472 & 0.160 & 0.498 & 0.350 \\
\midrule
\textit{Free-Form Multi-Agent} \\
gpt-oss-20b & 0.562 & 0.337 & 0.373 & 0.121 & 0.488 & 0.160 & 0.548 & 0.431 \\
Llama-3.1-8B-Instruct & 0.594 & 0.391 & 0.608 & \cellcolor{second} 0.326 & 0.580 & 0.271 & 0.466 & 0.354 \\
Llama-3.3-70B-Instruct & \cellcolor{second} 0.830 & \cellcolor{second} 0.604 & 0.730 & 0.312 & \cellcolor{best} 0.866 & 0.421 & 0.647 & \cellcolor{second} 0.498 \\
Meditron-70B & 0.582 & 0.437 & 0.473 & 0.215 & 0.501 & 0.206 & 0.548 & 0.422 \\
\midrule
\textit{Vital Trace} \\
Llama-3.1-8B-Instruct & 0.616 & 0.443 & \cellcolor{second} 0.793 & 0.224 & 0.648 & 0.225 & 0.568 & 0.391 \\
Llama-3.3-70B-Instruct & \cellcolor{best} 0.837 & \cellcolor{best} 0.622 & \cellcolor{best} 0.855 & \cellcolor{best} 0.345 & \cellcolor{second} 0.863 & \cellcolor{second} 0.436 & \cellcolor{best} 0.781 & \cellcolor{best} 0.603 \\
gpt-oss-20b & 0.622 & 0.397 & 0.473 & 0.201 & 0.692 & 0.289 & 0.628 & 0.344 \\
Qwen3-32B & 0.662 & 0.423 & 0.498 & 0.226 & 0.682 & 0.261 & 0.648 & 0.441 \\
Mixtral-8x7B-Instruct-v0.1 & 0.604 & 0.393 & 0.439 & 0.192 & 0.686 & \cellcolor{best} 0.485 & 0.534 & 0.366 \\
Meditron-70B & 0.647 & 0.458 & 0.619 & 0.244 & 0.643 & 0.186 & 0.543 & 0.374 \\
ClinicalCamel-70B & 0.670 & 0.373 & 0.493 & 0.223 & 0.673 & 0.216 & \cellcolor{second} 0.665 & 0.457 \\
\bottomrule
\end{tabular}
}
\label{tab:per_target_mimic}
\end{table*}

\begin{table*}[!ht]
\centering
\caption{Per-target predictive performance on eICU.}
\small
\resizebox{\textwidth}{!}{
\begin{tabular}{l|cc|cc|cc|cc}
\toprule
\multirow{2}{*}{Method} &
\multicolumn{2}{c|}{Vasopressor} &
\multicolumn{2}{c|}{Resp.\ support} &
\multicolumn{2}{c|}{Renal support} &
\multicolumn{2}{c}{Deterioration} \\
& AUROC $\uparrow$ & AUPRC $\uparrow$
& AUROC $\uparrow$ & AUPRC $\uparrow$
& AUROC $\uparrow$ & AUPRC $\uparrow$
& AUROC $\uparrow$ & AUPRC $\uparrow$ \\
\midrule
\textit{Single-Agent LLM} \\
gpt-oss-20b & 0.471 & 0.348 & 0.493 & 0.336 & 0.412 & 0.112 & 0.497 & 0.398 \\
Llama-3.1-8B-Instruct & 0.444 & 0.261 & 0.501 & 0.337 & 0.429 & 0.114 & 0.471 & 0.383 \\
Llama-3.3-70B-Instruct & 0.452 & 0.343 & 0.512 & 0.349 & 0.479 & 0.310 & 0.506 & 0.402 \\
Meditron-70B & 0.488 & 0.354 & 0.506 & 0.348 & 0.467 & 0.306 & 0.503 & 0.404 \\
\midrule
\textit{Free-Form Multi-Agent} \\
gpt-oss-20b & 0.521 & 0.371 & 0.547 & 0.396 & 0.471 & 0.341 & 0.528 & 0.416 \\
Llama-3.1-8B-Instruct & 0.538 & 0.376 & 0.593 & 0.438 & \cellcolor{second} 0.482 & \cellcolor{second} 0.352 & 0.534 & 0.421 \\
Llama-3.3-70B-Instruct & 0.546 & 0.382 & 0.568 & 0.425 & \cellcolor{best} 0.485 & \cellcolor{best} 0.355 & \cellcolor{second} 0.541 & \cellcolor{best} 0.425 \\
Meditron-70B & 0.517 & 0.368 & 0.552 & 0.401 & 0.463 & 0.336 & 0.523 & 0.414 \\
\midrule
\textit{Vital Trace} \\
Llama-3.1-8B-Instruct & 0.482 & 0.362 & \cellcolor{second} 0.679 & 0.420 & 0.453 & 0.307 & 0.500 & 0.381 \\
Llama-3.3-70B-Instruct & \cellcolor{best} 0.753 & \cellcolor{best} 0.415 & \cellcolor{best} 0.733 & 0.418 & 0.449 & 0.315 & \cellcolor{best} 0.612 & 0.422 \\
gpt-oss-20b & 0.536 & 0.381 & 0.612 & 0.472 & 0.468 & 0.328 & 0.519 & 0.418 \\
Qwen3-32B & 0.548 & 0.384 & 0.641 & \cellcolor{best} 0.506 & 0.472 & 0.337 & 0.526 & 0.421 \\
Mixtral-8x7B-Instruct-v0.1 & 0.527 & 0.373 & 0.601 & 0.461 & 0.458 & 0.319 & 0.514 & 0.412 \\
Meditron-70B & \cellcolor{second} 0.563 & \cellcolor{second} 0.389 & 0.638 & 0.495 & 0.481 & 0.324 & 0.531 & 0.417 \\
ClinicalCamel-70B & 0.545 & 0.383 & 0.629 & \cellcolor{second} 0.497 & 0.474 & 0.333 & 0.529 & \cellcolor{second} 0.423 \\
\bottomrule
\end{tabular}
}
\label{tab:per_target_eicu}
\end{table*}

Table~\ref{tab:secondary_metrics} reports additional evaluation metrics on MIMIC-IV and eICU, including probability calibration, early-warning utility, counterfactual consistency, and protocol safety indicators. Across most backbone models, Vital Trace consistently improves counterfactual directional consistency while reducing protocol-violation and auditor-failure rates relative to free-form multi-agent baselines. The strongest counterfactual consistency on MIMIC-IV is achieved by Vital Trace with Llama-3.3-70B-Instruct (\(0.852\)), while ClinicalCamel-70B attains the highest consistency on eICU (\(0.844\)). Vital Trace additionally achieves the lowest calibration error and protocol-violation rates across most evaluated settings, indicating improved longitudinal reasoning reliability under protocol-grounded communication.

Early-warning behavior exhibits substantial variability across datasets and backbone models. On MIMIC-IV, Vital Trace variants generally maintain stronger early-warning recall than single-agent baselines while preserving improved calibration. On eICU, several Vital Trace configurations achieve substantially longer lead times prior to intervention onset, including ClinicalCamel-70B (\(3.187\)h) and Meditron-70B (\(3.006\)h), although gains are less consistent across all backbone families due to higher intervention prevalence and denser intervention trajectories.

Tables~\ref{tab:per_target_mimic} and~\ref{tab:per_target_eicu} provide full per-target predictive performance on MIMIC-IV and eICU, respectively. On MIMIC-IV, Vital Trace with Llama-3.3-70B-Instruct achieves the strongest overall performance on vasopressor, respiratory-support, and generalized deterioration prediction, while remaining highly competitive on renal-support prediction. The largest gains are observed for respiratory-support prediction, where AUROC improves from \(0.730\) under free-form multi-agent reasoning to \(0.855\), suggesting that persistent patient-state evolution particularly benefits longitudinal respiratory instability modeling.

Results on eICU show a similar but weaker trend. Vital Trace consistently improves respiratory-support and deterioration prediction across most backbone families, while gains for renal-support prediction remain modest across all methods. Overall, the per-target analyses support the hypothesis that structured patient-state evolution is most beneficial for tasks requiring persistent temporal reasoning over evolving physiological instability.

\paragraph{Comparison with RETAIN-Style Sequential Modeling.}
We additionally evaluated a RETAIN-style sequential prediction baseline trained under a conventional supervised setting using a \(70\%/15\%/15\%\) train/validation/test split. The RETAIN-style model achieved Macro-AUROC/Macro-AUPRC of \(0.678/0.175\) on MIMIC-IV and \(0.652/0.360\) on eICU. These results are not directly comparable to the primary experiments because RETAIN is fully optimized through supervised training on the target datasets, whereas Vital Trace operates through frozen inference-time reasoning without gradient-based adaptation. Nevertheless, the comparison provides useful context: the strongest Vital Trace configuration exceeds RETAIN on MIMIC-IV (\(0.834/0.502\)), while remaining competitive on eICU, where the RETAIN-style baseline achieves slightly stronger ranking performance than several smaller LLM configurations. This difference is expected given the substantially higher intervention prevalence and denser trajectory structure in eICU, which favor supervised sequential optimization.

\end{document}